\documentclass[sigconf]{acmart}
\AtBeginDocument{%
  }

\setcopyright{cc}
\setcctype{by-nc-nd}
\copyrightyear{2026}
\acmYear{2026}
\acmConference[MM '26]{Proceedings of the 35th ACM International Conference on Multimedia}{November 10--14, 2026}{Rio de Janeiro, Brazil}
\acmBooktitle{Proceedings of the 35th ACM International Conference on Multimedia (MM '26), November 10--14, 2026, Rio de Janeiro, Brazil}
\acmDOI{10.1145/3767308.3836356}
\acmISBN{979-8-4007-2213-4/2026/11}

\usepackage{enumitem}
\usepackage{makecell} 

\usepackage{xcolor}

\begin{document}

\newcommand{\Name}{FreeLit: Paired-Free Indoor Relighting via Physics-Guided Diffusion}
\newcommand{\ShortName}{FreeLit}
\title[\ShortName]{\Name}

\author{Chi-En Yen}
\orcid{0009-0001-6922-9191}
\affiliation{%
  \institution{National Yang Ming Chiao Tung University}
  \city{Hsinchu}
  \country{Taiwan}
}
\email{miayan.cs13@nycu.edu.tw}

\author{Duy-Khanh Ngo}
\orcid{0009-0003-9311-9835}
\affiliation{%
  \institution{National Yang Ming Chiao Tung University}
  \city{Hsinchu}
  \country{Taiwan}}
\email{ngoduykhanh2009.ee13@nycu.edu.tw}

\author{Wen-Wei Tang}
\orcid{0009-0007-6025-3367}
\affiliation{%
  \institution{National Yang Ming Chiao Tung University}
  \city{Hsinchu}
  \country{Taiwan}}
\email{wei.cs14@nycu.edu.tw}

\author{Huu-Phu Do}
\orcid{0009-0006-7327-9016}
\affiliation{%
  \institution{National Yang Ming Chiao Tung University}
  \city{Hsinchu}
  \country{Taiwan}}
\email{dohuuphu25.ee11@nycu.edu.tw}

\author{Wen-Hsiao Peng}
\orcid{0000-0002-4421-8031}
\affiliation{%
  \institution{National Yang Ming Chiao Tung University}
  \city{Hsinchu}
  \country{Taiwan}}
\email{wpeng@cs.nctu.edu.tw}

\author{Ching-Chun Huang}
\orcid{0000-0002-4382-5083}
\affiliation{%
  \institution{National Yang Ming Chiao Tung University}
  \city{Hsinchu}
  \country{Taiwan}
  }
\email{chingchun@nycu.edu.tw}
\authornote{Ching-Chun Huang is the corresponding author.}




\begin{abstract}


Image-based indoor scene relighting remains challenging due to the complex interplay between cluttered geometry and local illumination, requiring precise modeling of light position, color, and intensity. 
Existing data-driven methods implicitly learn this relationship via paired multi-illumination datasets. Nevertheless, this data is costly and fails to scale, which is essential for accurate light-source-level control.
Conversely, inverse-rendering methods reduce the data dependency by incorporating physical priors; however, they lack the robustness of intrinsic estimation in challenging conditions.
In this paper, we present \ShortName, a paired-free framework for controllable indoor relighting that explicitly manipulates light-source location, color, and intensity. Instead of relying on paired supervision, we construct a physics-guided illumination prior from intrinsic scene properties, generating a structured lightmap along with a pseudo-relit image to guide diffusion-based synthesis. To address instability in intrinsic estimation, especially in low-light scenes, we introduce a relighting-guided intrinsic stabilization strategy that enforces illumination-invariant reflectance through structure-aware distillation and consistency constraints. Furthermore, we propose controllability-oriented evaluation metrics to quantify alignment with user-specified illumination color and intensity. Experimental results demonstrate that FreeLit achieves stable, physically consistent, and controllable relighting, with improved robustness in low-light indoor scenes, without requiring paired supervision. Our project page is available at \url{https://miayan0110.github.io/freelit.github.io/}.

\end{abstract}

\ccsdesc[500]{Computing methodologies~Computer vision}
\ccsdesc[500]{Computing methodologies~Image processing}

\keywords{Controllable Relighting, Unpaired Learning, Physics-Guided Learning}
\begin{teaserfigure}
  \centering
  \includegraphics[width=0.95\textwidth]{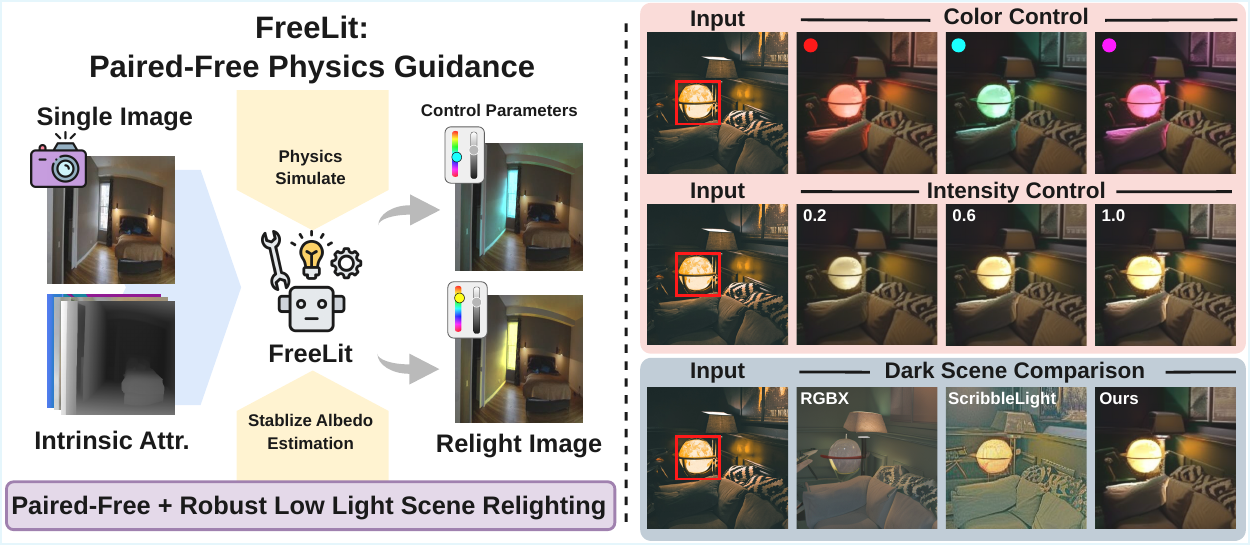}
  \caption{FreeLit enables paired-free controllable indoor relighting from a single image. The red box highlights the user-specified light source to be manipulated. Given this localized control, our method produces consistent color modulation and smooth intensity scaling while preserving scene geometry and shading. Compared with prior methods, FreeLit achieves more stable and realistic illumination, particularly in low-light indoor scenes, without requiring paired multi-illumination supervision.}
  \Description{teaser}
  \label{fig:teaser}
\end{teaserfigure}


\maketitle

\section{Introduction}
Indoor scene relighting aims to modify scene illumination while preserving geometric structure and semantic content. A key challenge in this problem is to enable \emph{light-source-level controllability}, where users can explicitly specify the location, color, and intensity of individual light sources. Such control is important for applications including virtual staging, interior design, film production, and augmented reality.

\textcolor{black}{Existing methods \cite{deng2024flashtex, choi2025scribblelight, bhattad2024stylitgan, magar2025lightlab, barron2014shape, narihira2015direct, li2020inverse} achieve high-fidelity relighting that relies on paired multi-illumination datasets, where the same scene is captured under varying lighting conditions for training.
Text-guided image relighting methods
~\cite{deng2024flashtex, bhattad2024stylitgan}
deliver photorealistic relighting results by leveraging strong generative priors. However, they fail to provide fine-grained control.
Alternatively, inverse-rendering methods ~\cite{barron2014shape,narihira2015direct,li2020inverse} leverage the Rendering Equation \cite{kajiya1986rendering} - which models surface appearance as lighting-material interaction - as a physical prior to decompose the RGB image to physically meaningful intrinsics, such as albedo, normal, and depth.
However, these methods often suffer from unstable albedo estimation in extreme lighting conditions — particularly in low-light indoor scenes, where reflectance and illumination are strongly entangled, leading to false relighting results. Recent methods 
\cite{choi2025scribblelight, magar2025lightlab} allow providing light-source-level control to the model; however, they require large-scale paired data, which remains scarce due to the specialized capture requirements. \textbf{This exposes a structural bottleneck: current models either offer visual quality without physical control, or provide controllability at the cost of data-driven instability in challenging scenes}.
}

In this paper, we present \ShortName, a paired-free framework for controllable indoor relighting from a single image. Our approach combines physics-guided illumination modeling with diffusion-based image synthesis to achieve both structured control and photorealistic results without requiring expensive paired datasets. 
Specifically, we proposed using a simplified physics-based formulation to estimate a structured lightmap from intrinsic properties, providing explicit spatial guidance to the Diffusion model for accurate relighting. Additionally, using forward rendering based on this formulation, a pseudo-relit image can be obtained that acts as supervision during training, effectively bypassing the need for ground-truth paired data.
By incorporating these physics-based priors, our framework maintains interpretable control over light-source parameters while avoiding the complexity of a multi-illumination setting.


In extreme lighting conditions, particularly low-light scenes, existing inverse rendering methods often struggle to produce stable albedo~\cite{careagaColorful}. To address this, we leverage a knowledge distillation strategy to train a student albedo predictor that preserves the teacher-guided expertise. This training is guided by our Relighting consistency constraints, which serve as an albedo augmentation to improve the model's robustness.

Finally, we observe that conventional evaluation metrics such as PSNR and LPIPS do not adequately reflect controllability with respect to user-specified lighting parameters. Thus, we introduce controllability-oriented metrics that explicitly measure alignment between generated results and target illumination in terms of color and intensity. In summary, our contributions are as follows:
\begin{itemize}
\item \textbf{Paired-free controllable indoor relighting.} FreeLit enables light-source-level control of indoor illumination without requiring paired multi-illumination data.
\item \textbf{Physics-guided diffusion framework.} We introduce a structured illumination prior derived from intrinsic scene properties, enabling scalable training and diffusion guidance.
\item \textbf{Intrinsic stabilization for low-light robustness.} We propose a relighting-consistent intrinsic learning strategy that improves reflectance estimation under challenging lighting conditions.
\item \textbf{Controllability-oriented evaluation metrics.} We introduce Color Accuracy to quantify alignment with user-specified illumination color.
\end{itemize}


\section{Related Works}
\subsection{Image-Based Relighting}
Image-based relighting aims to modify illumination in photographs without requiring full 3D reconstruction. The problem has been studied across multiple domains, including isolated objects \cite{deng2024flashtex,gao2024relightable,jin2024neural,poirier2024diffusion}, human portraits \cite{kim2024switchlight,hou2022face,mei2023lightpainter}, outdoor scenes \cite{liu2020learning,yu2020self,kocsis2024lightit}, and indoor environments \cite{xing2025luminet,bhattad2024stylitgan}. These works explore various strategies to manipulate lighting effects while preserving scene structure.

Different representations have been proposed to control illumination, ranging from explicit lighting models and geometric priors to latent representations learned through adversarial or diffusion-based frameworks. While object- and portrait-level relighting often benefit from more constrained geometry and lighting configurations, indoor scene relighting remains more challenging due to the increased complexity of spatial illumination and scene layout.

In this work, we focus on indoor scene relighting with precise light-source-level control, where users specify light-source masks, RGB color, and continuous intensity for each light source. This setting introduces challenges beyond global illumination transfer, requiring structured modeling of localized light interactions in cluttered environments.

\begin{figure*}[t]
  \centering
  \includegraphics[width=0.8\textwidth]{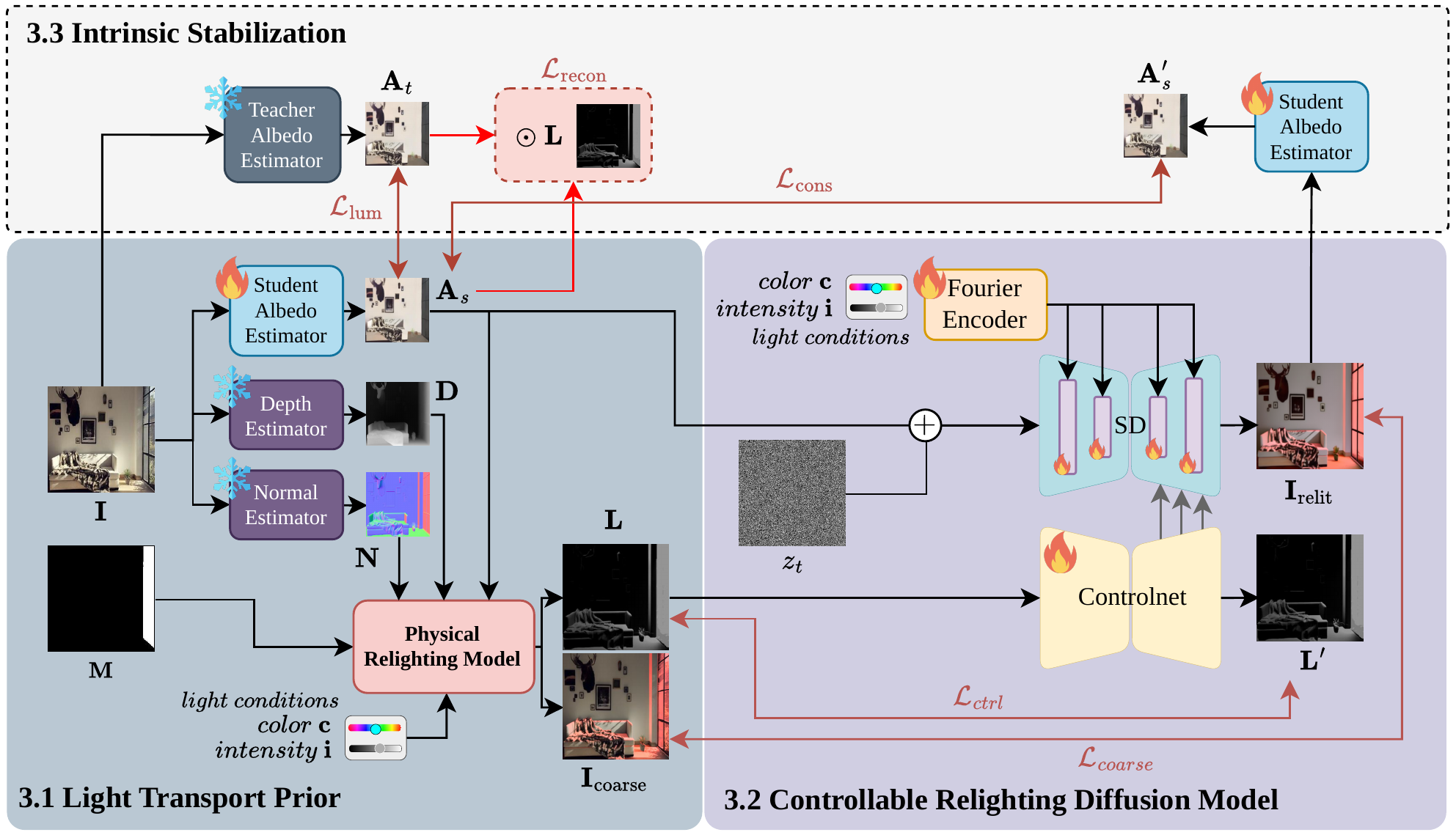}
  \caption{Overview of the proposed framework. Given an input image and user-specified lighting parameters (mask, color, and intensity), we construct a physics-guided lightmap to condition diffusion-based relighting. A pseudo-relit image is used as additional guidance during training. At inference, the model takes only the input image, lightmap, and user parameters to produce a relit result. The intrinsic stabilization module further improves robustness, especially in low-light scenes.
  }
  \Description{method_overview}
  \label{fig:method_overview}
\end{figure*}
\subsection{Indoor Scene Relighting}
Early works in indoor scene relighting focused on controlled domains, such as portraits or objects \cite{zhou2019deep,sun2019single,pandey2021total}, whereas recent efforts extend relighting to complex indoor environments with spatially distributed light sources and geometric interactions. Large-scale systems leverage paired multi-illumination capture to achieve high realism and light-source-level control \cite{magar2025lightlab}, but require costly, non-public, and difficult-to-scale paired supervision.

Beyond strictly paired supervision, few works explore latent-space or generative approaches for indoor relighting. \cite{bhattad2024stylitgan} performs relighting via latent control using adversarial training, \cite{xing2025luminet} integrates intrinsic representations with diffusion models for indoor scene relighting, while \cite{zhang2024latent} investigates emergent intrinsic structures learned through relighting objectives. Although these methods demonstrate promising visual quality, they do not provide explicit mask-based control with continuous color and intensity parameters. In contrast, our work explicitly targets paired-free indoor relighting with structured light-source-level control.

\subsection{Intrinsic Decomposition and Inverse Rendering}
Intrinsic image decomposition separates an image into an albedo and a shading component \cite{barron2014shape,narihira2015direct,careagaColorful}, providing structured representations for relighting. Neural inverse rendering methods further estimate scene geometry, reflectance, and illumination from a single image \cite{li2020inverse,taniai2018neural,zhu2022learning}. Several relighting approaches leverage these intrinsic and geometric components to manipulate illumination while preserving scene structure \cite{zeng2024rgb,kim2024switchlight}. However, their performance heavily depends on the quality of intrinsic estimation, particularly albedo. In complex indoor scenes, especially under low-light conditions, reflectance and illumination are strongly entangled, leading to unstable or inconsistent albedo predictions. To address this limitation, we introduce an intrinsic stabilization mechanism that improves albedo consistency under challenging illumination. Instead of treating intrinsic decomposition as a fixed preprocessing step, our method integrates relighting-driven constraints to refine albedo estimation, resulting in more coherent relighting.



\section{Method}


Controllable relighting of indoor scenes from a single image requires both geometric consistency and realistic appearance synthesis. Our framework addresses this challenge by combining a physics-guided illumination module with a diffusion-based generator. Given an input image and user-defined lighting parameters, including a light source mask, RGB color, and intensity, we first estimate intrinsic and geometric properties of the scene to construct a structured illumination representation, referred to as a \textit{lightmap}, along with a coarse relit image using a simplified physical model (Section~\ref{sec:physical_relighting}). The lightmap provides spatially structured guidance of illumination effects, enabling controllable relighting. We then leverage a fine-tuned diffusion model to refine the relit result, using both the lightmap and user-specified parameters as conditioning signals to produce photorealistic and controllable outputs (Section~\ref{sec:diffusion}). Finally, to address the instability of intrinsic estimation under challenging conditions such as low-light scenes, we introduce an intrinsic stabilization mechanism that improves albedo consistency during relighting (Section~\ref{sec:albedo_estimator}). The overall pipeline is illustrated in Figure~\ref{fig:method_overview}.

\subsection{Light Transport Prior}

\label{sec:physical_relighting}

%
To enable controllable relighting without paired multi-illumination data, we construct a structured illumination prior to guide the diffusion model. Specifically, we derive a lightmap $L$ that encodes the spatial distribution of illumination, along with a pseudo-relit image that provides coarse supervision.


While image formation is governed by the rendering equation~\cite{kajiya1986rendering}, solving it requires detailed geometry and material properties unavailable from a single image. We therefore adopt a Lambertian assumption~\cite{barron2014shape, li2020inverse} and approximate illumination using estimated surface normals and depth, capturing the dominant shading structure in a tractable manner. Higher-order effects such as cast shadows, specular reflections, and multi-bounce light are instead left to the diffusion model.

Given predicted surface normals $N$ and depth $D$, we approximate direct illumination as:
\begin{equation}
\label{eq:direct}
L_{\text{direct}}(x) =
\sum_{p \in \mathcal{S}(M)}
\frac{\max(0, N(x)\cdot \ell_{xp})}
{\|\ell_{xp}\|^2 + \epsilon},
\end{equation}
where $\ell_{xp}$ denotes the direction vector from pixel $x$ to a sampled light position $p$. This formulation captures the primary lighting effects through Lambertian cosine shading and distance-based attenuation following the inverse-square law.

While direct illumination models the primary light contribution, real-world scenes also exhibit indirect illumination due to multi-bounce light interactions. Instead of explicitly modeling these complex effects, we introduce a lightweight ambient approximation inspired by real-time rendering. We model environment illumination as a spatially uniform component:
\begin{equation}
L_{\text{env}} = \beta \, \mathbf{1},
\label{eq:ambient}
\end{equation}
where $\beta$ represents global background illumination intensity.

We further incorporate a diffuse approximation of indirect lighting by redistributing direct illumination:
\begin{equation}
\label{eq:coarse}
L_{\text{ambient}}(x) =
L_{\text{env}} + \alpha \, \mathbb{E}_x[L_{\text{direct}}(x)],
\end{equation}
where $\mathbb{E}_x[\cdot]$ denotes spatial averaging. We then combine the direct and ambient components to form the final lightmap:
\begin{equation}
\label{eq:lightmap}
L(x) = L_{\text{direct}}(x) + L_{\text{ambient}}(x).
\end{equation}
Using the estimated albedo $A$, we further construct a pseudo relit image:
\begin{equation}
I_{\text{coarse}}(x) = A(x) \odot (L(x) \odot c),
\end{equation}
where $\odot$ denotes element-wise multiplication and $c=(c_r,c_g,c_b)$ is the user-specified RGB light color.

\subsection{Controllable Relighting Diffusion Model}

\label{sec:diffusion}
To achieve controllable and photorealistic relighting, we leverage a diffusion-based framework conditioned on the structured cues derived in Section~\ref{sec:physical_relighting}. We build upon a pretrained Stable Diffusion (SD)~\cite{rombach2022high} model and perform lightweight fine-tuning on its cross-attention layers, preserving its strong generative prior while enabling controllable relighting.

Given user-specified lighting parameters, including color $c \in \mathbb{R}^3$ and intensity $i \in \mathbb{R}$, we map these low-dimensional inputs into a higher-dimensional embedding space using Fourier feature encoding. Specifically, for each scalar component $u \in \{i, c_r, c_g, c_b\}$, we compute:
\begin{equation}
\gamma(u) = \left[ \sin(2^k \pi u), \cos(2^k \pi u) \right]_{k=0}^{F-1}.
\end{equation}
where $F$ is the number of frequency bands. The encoded features are then projected through a lightweight MLP and injected into the UNet via cross-attention layers.

In addition to parameterized conditioning, we incorporate spatial guidance through ControlNet. The physics-guided lightmap $L$ is used as the conditioning signal, and we enforce its consistency via an auxiliary reconstruction loss:
\begin{equation}
\mathcal{L}_{\text{ctrl}} = \| L' - L \|_1,
\end{equation}
where $L'$ denotes the ControlNet-predicted lightmap.

Furthermore, we introduce an image-space guidance term based on the pseudo-relit image to guide the diffusion process toward physically plausible illumination. Specifically, for a sampled timestep $t$, we concatenate the predicted albedo $A_s$ with the noisy latent $z_t$ as the input to the diffusion model. The model then predicts noise $\epsilon_\theta(\cdot)$ and reconstructs the clean image $\hat{x}_0$, which is decoded into the relit image $I_{\text{relit}}$. We compute:
\begin{equation}
\mathcal{L}_{\text{coarse}} = \| I_{\text{relit}} - I_{\text{coarse}} \|_1,
\end{equation}
which encourages the generated result to follow the structured illumination cues while maintaining photorealistic flexibility.

\subsection{Intrinsic Stabilization Strategy }
\label{sec:albedo_estimator}
Accurate albedo estimation is critical for reliable relighting, yet it becomes unstable in low-light indoor scenes due to the entanglement of reflectance and illumination. To improve robustness without requiring albedo supervision, we introduce an intrinsic stabilization strategy based on self-supervised constraints aligned with the relighting objective. We adopt a teacher-student formulation, where the teacher model remains frozen to preserve its pretrained intrinsic estimation, and the student model is adapted to the relighting task under additional constraints, enabling more stable albedo estimation.

Let $A_s$ and $A_t$ be the albedo predicted from the input image $I$ by the student and teacher ~\cite{careagaColorful} models respectively.
We first introduce a structure-preserving distillation signal by matching luminance rather than full RGB values:
\begin{equation}
\mathcal{L}_{\text{lum}} =
\big\| Y(A_s) - Y(A_t) \big\|_1,
\end{equation}
where $Y(\cdot)$ converts an RGB image into luminance. Focusing on luminance preserves reliable reflectance structure while reducing sensitivity to illumination-dependent color variations. We further apply a decayed weighting schedule to $\mathcal{L}_{\text{lum}}$ to prevent overfitting to teacher-specific bias.

We further enforce a physics-aligned reconstruction constraint using the simulated illumination from Section~\ref{sec:physical_relighting}. We first construct a relit image using the predicted albedo:
\begin{equation}
I_{\text{recon}} = A_s \odot L.
\end{equation}
To account for global illumination discrepancies, we perform a per-image affine alignment between the reconstructed image and the target:
\begin{equation}
\mathcal{L}_{\text{recon}} =
\big\| 
\left( s \cdot (A_s \odot L) + b \right)
- A_t \odot L 
\big\|_1,
\end{equation}
where $s$ and $b$ are scalar parameters estimated via least-squares fitting for each image. This alignment absorbs global intensity and bias differences, allowing the model to focus on reflectance consistency rather than exact brightness matching.


Finally, we introduce a relighting consistency constraint. Given a relit image $I_{\text{relit}}$ generated under a sampled lighting configuration, we re-estimate its albedo $A_s'$ using the same predictor. Since relighting modifies illumination but not surface reflectance, the predicted albedo should remain stable:
\begin{equation}
\mathcal{L}_{\text{cons}} =
\big\| A_s - A_s' \big\|_2^2.
\end{equation}

The overall albedo stabilization objective is defined as:
\begin{equation}
\mathcal{L}_{\text{intr}} =
\lambda_{\text{lum}} \mathcal{L}_{\text{lum}}
+ \lambda_{\text{recon}} \mathcal{L}_{\text{recon}}
+ \lambda_{\text{cons}} \mathcal{L}_{\text{cons}}.
\end{equation}
These complementary objectives improve the stability of albedo estimation under complex and low-light illumination, leading to more reliable physics-guided relighting.

\subsection{Training Objective}
The overall training objective jointly optimizes the diffusion relighting module and the albedo stabilization network:
\begin{equation}
\mathcal{L}_{\text{total}} =
\mathcal{L}_{\text{diff}} 
+ \lambda_{\text{ctrl}} \mathcal{L}_{\text{ctrl}}
+ \lambda_{\text{coarse}} \mathcal{L}_{\text{coarse}}
+ \lambda_{\text{intr}} \mathcal{L}_{\text{intr}},
\end{equation}
where $\mathcal{L}_{\text{diff}} = \| \epsilon - \epsilon_\theta(z_t, t, \cdot) \|_2^2$ denotes the standard diffusion denoising objective in latent space. This multi-objective formulation balances illumination guidance and generative realism, allowing the model to refine beyond the coarse physical approximation.

\begin{figure*}[t]
    \centering
    \includegraphics[width=0.85\linewidth]{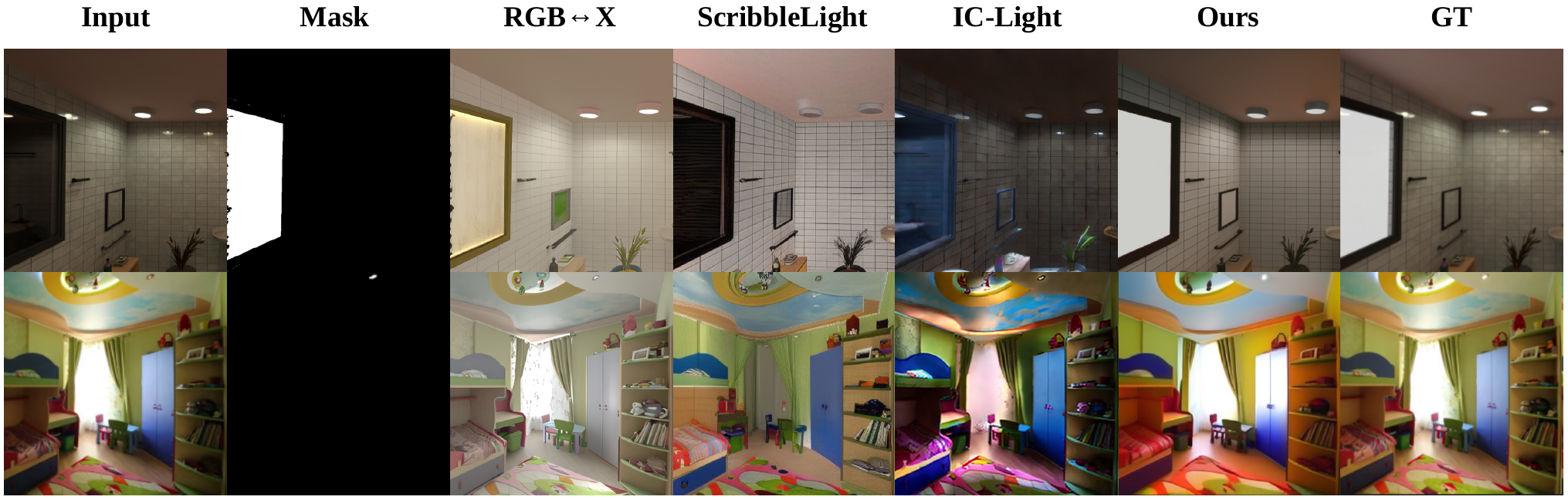}
    \caption{
    Qualitative relighting comparison against the ground truth~(GT). RGB$\leftrightarrow$X~\cite{zeng2024rgb} over-brightens and suppresses the ambient illumination, ScribbleLight~\cite{choi2025scribblelight} preserves scene content but barely relights, and IC-Light~\cite{zhang2025scaling} produces stylized, globally shifted illumination, whereas \ShortName follows the target most faithfully.
    }
    \Description{comparison}
    \label{fig:comparison}
\end{figure*}
\begin{figure*}[t]
    \centering
    \includegraphics[width=0.85\linewidth]{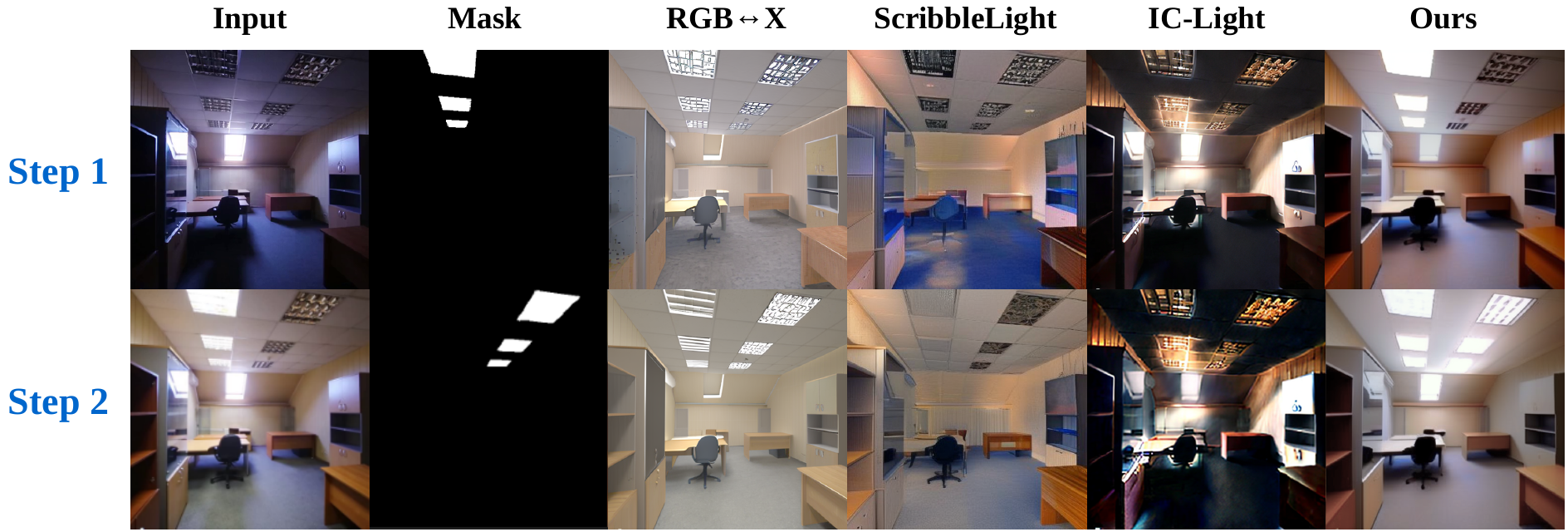}
    \caption{
    Step-by-step relighting under low-light scenes. Light sources are turned on sequentially: Step~2 takes the relit result of Step~1 as its input. From left to right: input, source mask, RGB$\leftrightarrow$X, ScribbleLight, IC-Light, and \ShortName (reference shown on the right). Baselines introduce haze and artifacts or fail to relight the specified source, whereas \ShortName preserves image content and accumulates the relighting effects across steps.
    }
    \Description{dark comparison}
    \label{fig:dark_comparison}
\end{figure*}

\section{Experiments}
\subsection{Experimental Setup}
We build our model upon Stable Diffusion v2.1~\cite{rombach2022high} and fine-tune its cross-attention layers. The model is trained on approximately 12K indoor images for 20 epochs on a single NVIDIA RTX A6000 GPU with a batch size of 1 and a constant learning rate of $5\times10^{-6}$. We employ the AdamW optimizer with a weight decay of $0.01$. Unless otherwise specified, all loss function hyperparameters are set to $1.0$ to maintain a balanced contribution from each objective.

\subsubsection{Dataset.}

We train our model on a subset of the LSUN Bedroom dataset~\cite{yu15lsun}, comprising approximately 12,000 indoor images utilized without paired supervision. For evaluation, we leverage the BigTime dataset~\cite{li2018learning} to facilitate full-reference assessment against ground-truth relighting results. To further evaluate performance across diverse lighting scenarios, we supplement our test set with synthetic data from Infinigen Indoors~\cite{infinigen2024indoors} and in-the-wild internet images. Detailed procedures for synthetic data generation and real-world image collection are provided \textbf{in the Supplementary Material.}


\subsubsection{Metrics.}
We employ PSNR, SSIM, and LPIPS to evaluate overall image fidelity against ground truth. However, quantifying chromatic control is inherently challenging due to the lack of ground-truth pairs for arbitrary lighting. To this end, we introduce two complementary non-reference metrics: color accuracy ($\text{CA}$) and its standard deviation ($\text{CA}_{\text{std}}$).



Color accuracy measures whether the chromatic change induced by relighting aligns with the target illumination color. Following prior work~\cite{magar2025lightlab}, we first estimate the illumination change by computing the pixel-wise difference between the relit image $I_{\text{relit}}$ and the input image $I$:
\begin{equation}
\Delta c(x) = I_{\text{relit}}(x) - I(x),
\end{equation}
where $x$ denotes a pixel location and the target illumination color is denoted by $c \in \mathbb{R}^3$.

We then measure the alignment between the observed color change and the target illumination color $c$ using a weighted cosine similarity:
\begin{equation}
\text{CA} = \frac{1}{\sum_x \|\Delta c(x)\|}
\sum_x \|\Delta c(x)\| \cdot
\frac{\Delta c(x) \cdot c}{\|\Delta c(x)\| \|c\|}.
\end{equation}
This formulation emphasizes pixels with stronger illumination changes while reducing the influence of negligible variations. Higher CA indicates that the relighting result better follows the desired color direction. To further evaluate the consistency of color control across different images, we additionally report the standard deviation of CA (CA$_{\text{std}}$), which measures the variation of color alignment performance across scenes. Lower CA$_{\text{std}}$ indicates more stable and consistent color alignment across diverse scenes.

For evaluating intensity control, we measure whether the brightness change follows the user-specified intensity levels. Given an input image $I$ and its relit result $I_{\text{relit}}$, we first compute the luminance at each pixel $x$:
\begin{equation}
Y(x) = 0.2126 R(x) + 0.7152 G(x) + 0.0722 B(x),
\end{equation}
where $x$ denotes a pixel location. We then define $\Omega_M$ as the user-specified light source mask region, and compute the mean luminance within this region:
\begin{equation}
\bar{Y} = \frac{1}{|\Omega_M|} \sum_{x \in \Omega_M} Y(x).
\end{equation}
Given a set of relit images with different target intensities $\{i_k\}$, we obtain the corresponding mean luminance values $\{\bar{Y}_k\}$ and evaluate their consistency with the target intensities using rank and linear correlation:
\begin{equation}
\rho_s = \text{Spearman}(\{i_k\}, \{\bar{Y}_k\}), \quad
\rho_p = \text{Pearson}(\{i_k\}, \{\bar{Y}_k\}).
\end{equation}
Spearman correlation measures whether the predicted brightness changes follow the correct ordering, while Pearson correlation evaluates linear consistency. A higher correlation indicates more accurate, monotonic intensity control.

To further support the effectiveness of the proposed controllability metrics, we conduct a user study to evaluate perceptual quality in terms of color fidelity and spatial control based on ranking. In addition, we conduct a separate user study on in-the-wild images focusing on perceptual realism. Details of the study protocol and the in-the-wild user study are provided in \textbf{the supplementary material}.

\subsubsection{Comparison methods.}
We compare \ShortName against three recent diffusion-based relighting approaches: RGB$\leftrightarrow$X~\cite{zeng2024rgb}, which performs relighting via intrinsic decomposition within a diffusion framework; ScribbleLight~\cite{choi2025scribblelight}, which enables spatial illumination control through user-provided masks; and IC-Light~\cite{zhang2025scaling}, a large-scale text-conditioned relighting model. We use two evaluation settings: (i) paired relighting in the on/off configuration, as existing methods primarily support binary illumination changes without explicit intensity control, and (ii) color controllability. Because IC-Light is driven by text prompts, we evaluate it following its paper's protocol; ScribbleLight is evaluated only under the paired setting, as it does not support explicit control over illumination color. In all comparisons, methods are provided with the same light-source mask where applicable.

\subsection{Comparison with State-of-the-Art}
Table~\ref{tab:quantitative_main} shows that our method achieves the best performance across all metrics, as well as the highest realism score in the user study, indicating that physically structured supervision combined with diffusion-based generation can produce faithful relighting without relying on paired multi-illumination training data. As shown in Figure~\ref{fig:comparison}, RGB$\leftrightarrow$X preserves scene structure relatively well, but tends to over-brighten the image and suppress the original ambient illumination, resulting in less realistic lighting. In contrast, ScribbleLight struggles to preserve image content, often introducing noticeable artifacts and distortions.

Figure~\ref{fig:dark_comparison} further shows sequential, per-light-source relighting, where light sources are turned on one at a time and each step relights the output of the previous one. RGB$\leftrightarrow$X again over-brightens the scene, while ScribbleLight produces noticeable artifacts and fails to preserve image content. In contrast, our method maintains stable illumination and clean details while consistently accumulating the relighting effects across steps.

\begin{table}[t]
\centering
\caption{Quantitative comparison of relighting performance. We evaluate on two datasets, focusing on binary light-source states (on/off), together with a realism user study (average rank score, higher is better). \colorbox{red!15}{Pink} indicates the best performance.}
\resizebox{\linewidth}{!}{
\begin{tabular}{lccc|ccc|c}
\toprule
& \multicolumn{3}{c}{Synthetic~\cite{infinigen2024indoors}} & \multicolumn{3}{c}{BigTime~\cite{li2018learning}} & Realism \\
\cmidrule(lr){2-4} \cmidrule(lr){5-7} \cmidrule(lr){8-8}
Method & PSNR$\uparrow$ & SSIM$\uparrow$ & LPIPS$\downarrow$ & PSNR$\uparrow$ & SSIM$\uparrow$ & LPIPS$\downarrow$ & User Study$\uparrow$ \\
\midrule
RGB$\leftrightarrow$X~\cite{zeng2024rgb} & 10.11 & 0.4475 & 0.4625 & 11.44 & 0.5331 & 0.4275 & 1.59 \\
ScribbleLight~\cite{choi2025scribblelight} & 13.22 & 0.4323 & 0.4504 & 13.60 & 0.4935 & 0.4848 & 2.08 \\
IC-Light~\cite{zhang2025scaling} & 13.65 & 0.5197 & 0.4127 & 12.95 & 0.5317 & 0.3984 & 2.68 \\
Ours & \colorbox{red!15}{15.10} & \colorbox{red!15}{0.5927} & \colorbox{red!15}{0.3931} & \colorbox{red!15}{15.55} & \colorbox{red!15}{0.6612} & \colorbox{red!15}{0.3821} & \colorbox{red!15}{3.21} \\
\bottomrule
\end{tabular}
}
\label{tab:quantitative_main}
\end{table}

\begin{table}[ht]
\centering
\caption{Quantitative comparison on color controllability. We report Color Accuracy (CA) across different illumination colors, its standard deviation (CA$_{\text{std}}$), and user study scores. Higher CA and lower CA$_{\text{std}}$ indicate better accuracy and consistency of color control, while higher user study scores reflect better perceptual quality in terms of color correctness and spatial control.}
\begin{tabular}{lccc}
\toprule
Method & CA$\uparrow$ & CA\_std$\downarrow$ & $\text{User Study}\uparrow$ \\
\midrule
RGB$\leftrightarrow$X~\cite{zeng2024rgb} & 0.6211 & 0.4544 & 1.42 \\
IC-Light~\cite{zhang2025scaling} & 0.6859 & 0.4916 & 1.81 \\
Ours & {\colorbox{red!15}{0.9295}} & {\colorbox{red!15}{0.1062}} & {\colorbox{red!15}{2.65}} \\
\bottomrule
\end{tabular}
\label{tab:color_metric}
\end{table}

\subsection{Controllability Evaluation}

\subsubsection{Color controllability.}
\begin{figure*}[!t]
    \centering
    \includegraphics[width=0.85\textwidth]{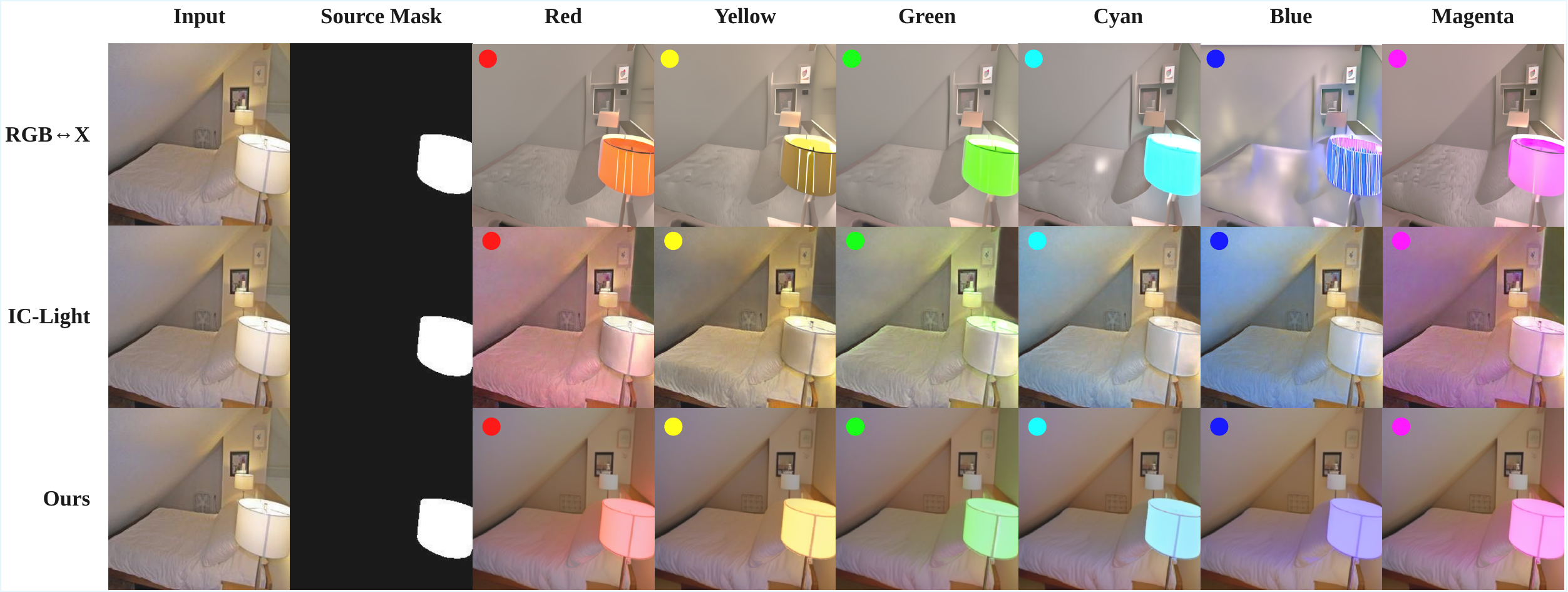}
    \caption{
    Color controllability under varying illumination colors with fixed intensity. Our method produces consistent and spatially coherent chromatic changes that follow scene geometry and shading, rather than applying simple global color transformations.
    }
    \Description{color_demo}
    \label{fig:color_demo}
\end{figure*}

As shown in Table~\ref{tab:color_metric}, our method achieves the highest color accuracy (CA) while significantly reducing CA$_{\text{std}}$, indicating both accurate and stable color alignment across different scenes. In contrast, although IC-Light attains relatively high CA, it exhibits a much larger CA$_{\text{std}}$, suggesting inconsistent behavior under varying conditions. RGB$\leftrightarrow$X performs worse on both metrics, indicating limited ability to accurately track the target illumination color.
As illustrated in Figure~\ref{fig:color_demo}, RGB$\leftrightarrow$X tends to produce over-brightened results with poor color fidelity, while IC-Light often introduces global color shifts that disregard scene geometry. In contrast, our method yields spatially coherent chromatic changes that remain consistent across target colors, resulting in more realistic and controllable relighting. The user study further confirms the superior perceptual quality of our method.

\subsubsection{Intensity controllability.}
\begin{figure}[ht]
    \centering
    \includegraphics[width=0.9\linewidth]{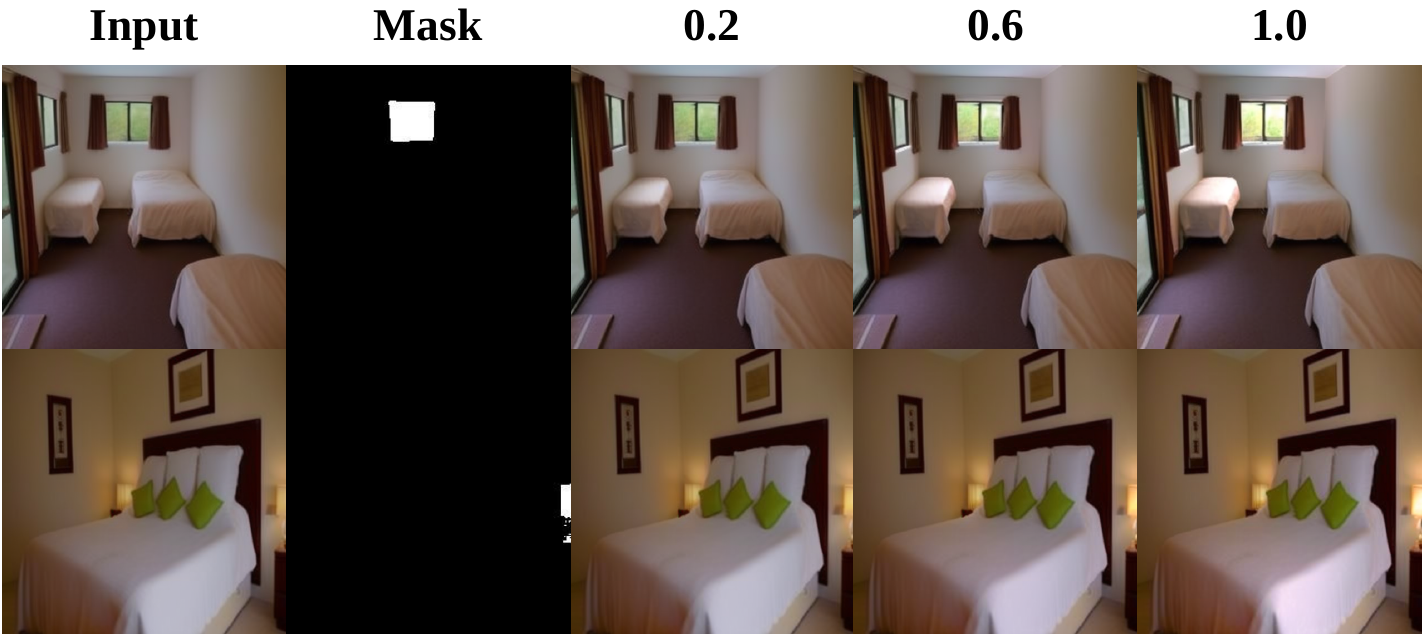}
    \caption{
    Qualitative results of intensity controllability under fixed illumination color. As the input intensity increases, the generated images exhibit smooth and consistent brightness variations while preserving object boundaries and shading.
    }
    \Description{intensity_demo}
    \label{fig:intensity_demo}
\end{figure}

We evaluate intensity control by measuring the correlation between the user-specified intensity parameter and the resulting image luminance. On the whole image, our method achieves a Spearman correlation of $0.8857$ and a Pearson correlation of $0.6895$, indicating that increasing the target intensity reliably yields brighter relit images in a near-monotonic fashion. When restricting evaluation to the masked light region, the Pearson correlation further improves to $0.8001$ while the Spearman score remains high, reflecting stronger localized intensity control.
Figure~\ref{fig:intensity_demo} shows qualitative results under varying intensity levels. As the input intensity increases, the generated images exhibit smooth, consistent brightness changes while preserving object boundaries and scene structure, demonstrating reliable intensity control.

\subsection{Ablation Study}
\label{sec:ablation}
To evaluate the importance of our proposed components, we present quantitative results for 4 ablation versions in Table~\ref{tab:ablation_full}, corresponding to the 4 settings below. \textbf{More qualitative results are provided in the supplementary}.



\subsubsection{(A) Pseudo Relighting Baseline.} 

We first examine the performance of the coarse relit image $I_{coarse}$ generated exclusively via the physical formulation in Equation \ref{eq:coarse}. As shown in Table~\ref{tab:ablation_full}, despite being a physically accurate model, this baseline achieves the lowest performance. Although the method produces structurally plausible results, reliance on pre-trained intrinsic models often leads to overly smooth geometry and a loss of high-frequency details, resulting in a lack of photorealism.

\subsubsection{(B) Diffusion with Intrinsic Guidance.} 
We next introduce the diffusion model with intrinsic conditioning, including predicted albedo, geometry, and mask. As shown in Table~\ref{tab:ablation_full}, this setting significantly improves SSIM compared to Setting A, indicating better perceptual quality and structural fidelity. However, PSNR improvement remains limited, and LPIPS is slightly degraded. This suggests that while diffusion enhances visual realism, the lack of structured illumination guidance results in inconsistent light distribution. 


\subsubsection{(C) Lightmap-based Conditioning.}
We further replace the mask-based conditioning with the proposed lightmap, which provides a structured representation of illumination. 
By explicitly modeling light transport, this setting reduces ambiguity in lighting distribution and improves relighting accuracy, improving the overall performance compared to setting B.

\subsubsection{(D) Full Model with Intrinsic Stabilization.}
\begin{figure}[ht]
    \centering
    \includegraphics[width=\linewidth]{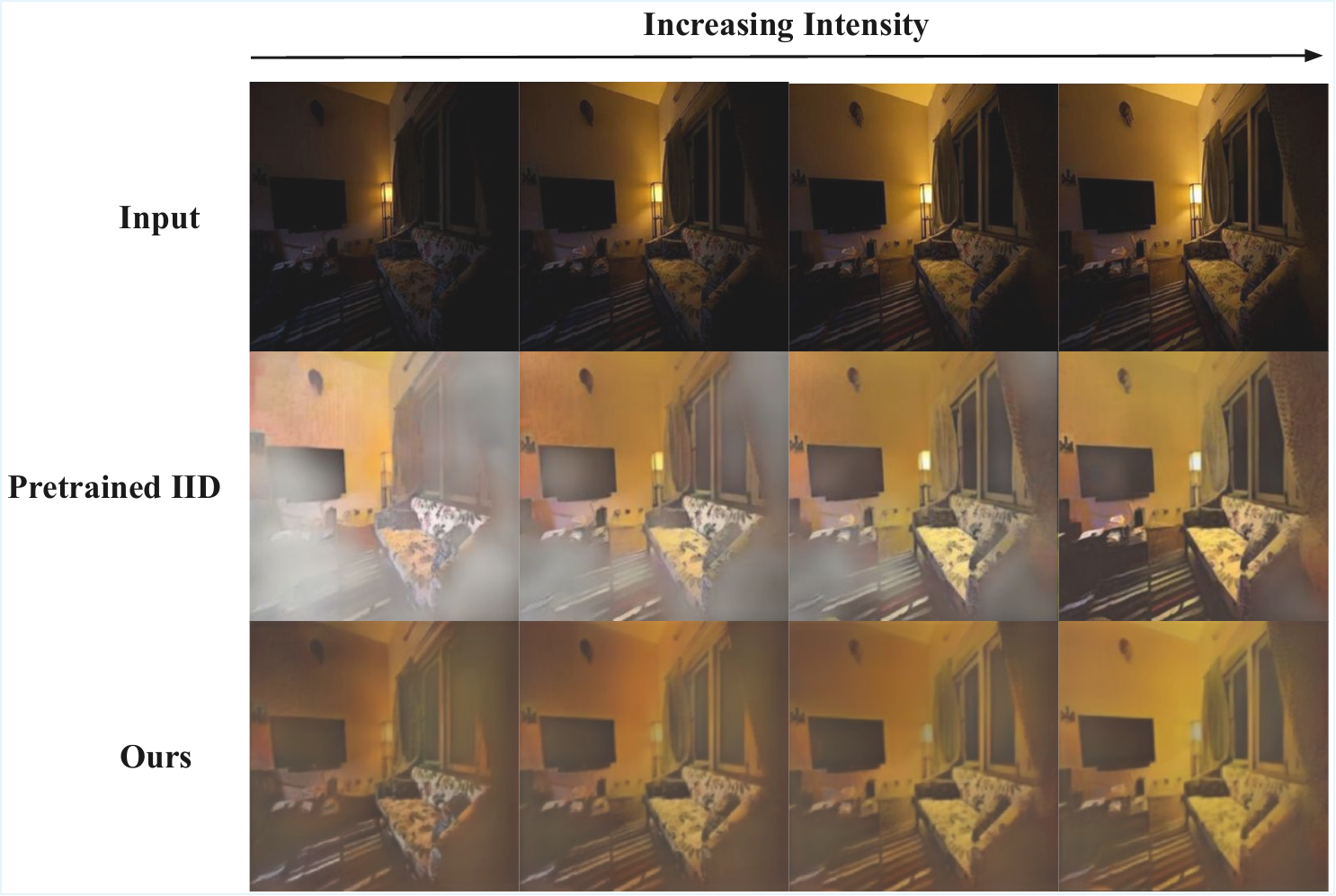}
    \caption{
    Ablation on albedo estimation under low-light conditions. We vary the illumination intensity for the same scene while examining the consistency of the predicted albedo.
    Compared to the pretrained teacher model~\cite{careagaColorful}, which exhibits noticeable variations under different lighting, our method produces more stable and illumination-invariant albedo, maintaining consistent reflectance across all conditions.
    }
    \Description{ablation_albedo}
    \label{fig:ablation_albedo}
\end{figure}

Finally, we incorporate the proposed intrinsic stabilization strategy by replacing the teacher albedo with the student-predicted albedo. 
Table~\ref{tab:ablation_full} shows that this setting achieves the best overall performance, with the highest PSNR and SSIM. 
As can be seen in Figure \ref{fig:ablation_albedo}, the teacher model produces blurry albedo in low-light cases, while our proposed method provide clean albedo.
This improvement demonstrates that stable and illumination-invariant albedo estimation is critical for reliable relighting. 
In particular, intrinsic stabilization reduces errors under challenging conditions such as low-light scenes, leading to more consistent reflectance estimation and improved overall fidelity.

\begin{table}[t]
\centering
\caption{Ablation study of different components in our framework. 
We progressively add intrinsic cues, physics guidance, diffusion modeling (DM), and intrinsic stabilization.}
\resizebox{\linewidth}{!}{
\begin{tabular}{lcccccc} 
\toprule
Setting & \makecell[c]{Light Transport \\ Prior} & \makecell[c]{Relighting \\ Diffusion Model} & \makecell[c]{Intrinsic \\ Stabilization} & PSNR$\uparrow$ & SSIM$\uparrow$ & LPIPS$\downarrow$ \\
\midrule
A  & \checkmark &  &  & 14.77 & 0.4771 & 0.3761 \\
B &  & \checkmark &  & 14.79 & 0.6304 & 0.3848 \\
C  & \checkmark & \checkmark &  & 14.90 & 0.6376 & 0.3846 \\
D  & \checkmark & \checkmark & \checkmark & {\colorbox{red!15}{15.55}} & {\colorbox{red!15}{0.6612}} & {\colorbox{red!15}{0.3821}} \\
\bottomrule
\end{tabular}
}
\label{tab:ablation_full}
\end{table}



\section{Conclusion}
We presented, \ShortName, a controllable indoor relighting framework that enables light-source-level manipulation of illumination color and intensity without relying on \textbf{paired} multi-illumination training data. Instead of collecting large-scale real-world paired supervision, we leverage physics-based illumination simulation to provide structured lighting priors, and integrate them with diffusion-based image generation for perceptually realistic relighting. Our experiments demonstrate that physically structured conditioning yields competitive quantitative performance while maintaining stable geometric and chromatic behavior. The proposed controllability metrics further reveal that our model achieves consistent intensity scaling and directional color alignment at the light-source level. Through ablation analysis, we show that diffusion-based generation enhances perceptual coherence beyond deterministic physical simulation, and that relighting-consistent intrinsic learning improves robustness under challenging illumination conditions. Overall, this work suggests that structured illumination modeling offers a viable alternative to data-intensive paired supervision for controllable relighting. We hope this direction encourages further exploration of physics-guided generative models for interpretable and user-driven scene editing.

\begin{acks}
This work was financially supported in part (project number: 112UA10019) by the Co-creation Platform of the Industry Academia Innovation School, NYCU, under the framework of the National Key Fields Industry-University Cooperation and Skilled Personnel Training Act, from the Ministry of Education (MOE) and industry partners in Taiwan. It also supported in part by the National Science and Technology Council, Taiwan, under Grant NSTC-115-2634-F-A49-011-, NSTC-114-2218-E-A49-024-, Grant NSTC-115-2425-H-A49-001, Grant NSTC-114-2622-E-A49-027, Grant NSTC-115-2221-E-A49 -124 -MY3, Grant NSTC-115-2218-E-A49 -017 and in part by the Higher Education Sprout Project of the National Yang Ming Chiao Tung University and the Ministry of Education (MOE), Taiwan. It is also partly supported by MediaTek Inc., Hon Hai Research Institute, and Industrial Technology Research Institute. 
\end{acks}

\clearpage
\bibliographystyle{ACM-Reference-Format}
\bibliography{main}

@inproceedings{magar2025lightlab,
  title={Lightlab: Controlling light sources in images with diffusion models},
  author={Magar, Nadav and Hertz, Amir and Tabellion, Eric and Pritch, Yael and Rav-Acha, Alex and Shamir, Ariel and Hoshen, Yedid},
  booktitle={Proceedings of the Special Interest Group on Computer Graphics and Interactive Techniques Conference Conference Papers},
  pages={1--11},
  year={2025}
}

@article{barron2014shape,
  title={Shape, illumination, and reflectance from shading},
  author={Barron, Jonathan T and Malik, Jitendra},
  journal={IEEE transactions on pattern analysis and machine intelligence},
  volume={37},
  number={8},
  pages={1670--1687},
  year={2014},
  publisher={IEEE}
}

@inproceedings{narihira2015direct,
  title={Direct intrinsics: Learning albedo-shading decomposition by convolutional regression},
  author={Narihira, Takuya and Maire, Michael and Yu, Stella X},
  booktitle={Proceedings of the IEEE international conference on computer vision},
  pages={2992--2992},
  year={2015}
}

@inproceedings{li2020inverse,
  title={Inverse rendering for complex indoor scenes: Shape, spatially-varying lighting and svbrdf from a single image},
  author={Li, Zhengqin and Shafiei, Mohammad and Ramamoorthi, Ravi and Sunkavalli, Kalyan and Chandraker, Manmohan},
  booktitle={Proceedings of the IEEE/CVF conference on computer vision and pattern recognition},
  pages={2475--2484},
  year={2020}
}

@inproceedings{rombach2022high,
  title={High-resolution image synthesis with latent diffusion models},
  author={Rombach, Robin and Blattmann, Andreas and Lorenz, Dominik and Esser, Patrick and Ommer, Bj{\"o}rn},
  booktitle={Proceedings of the IEEE/CVF conference on computer vision and pattern recognition},
  pages={10684--10695},
  year={2022}
}

@ARTICLE{careagaColorful,
 author={Chris Careaga and Ya\u{g}{\i}z Aksoy},
 title={Colorful Diffuse Intrinsic Image Decomposition in the Wild},
 journal={ACM Trans. Graph.},
 year={2024},
 volume = {43},
 number = {6},
 articleno = {178},
 numpages = {12},
}

@inproceedings{zhou2019deep,
  title={Deep single-image portrait relighting},
  author={Zhou, Hao and Hadap, Sunil and Sunkavalli, Kalyan and Jacobs, David W},
  booktitle={Proceedings of the IEEE/CVF international conference on computer vision},
  pages={7194--7202},
  year={2019}
}

@article{sun2019single,
  title={Single image portrait relighting.},
  author={Sun, Tiancheng and Barron, Jonathan T and Tsai, Yun-Ta and Xu, Zexiang and Yu, Xueming and Fyffe, Graham and Rhemann, Christoph and Busch, Jay and Debevec, Paul E and Ramamoorthi, Ravi},
  journal={ACM Trans. Graph.},
  volume={38},
  number={4},
  pages={79--1},
  year={2019}
}

@article{pandey2021total,
  title={Total relighting: learning to relight portraits for background replacement.},
  author={Pandey, Rohit and Orts-Escolano, Sergio and Legendre, Chloe and Haene, Christian and Bouaziz, Sofien and Rhemann, Christoph and Debevec, Paul E and Fanello, Sean Ryan},
  journal={ACM Trans. Graph.},
  volume={40},
  number={4},
  pages={43--1},
  year={2021}
}

@inproceedings{bhattad2024stylitgan,
  title={Stylitgan: Image-based relighting via latent control},
  author={Bhattad, Anand and Soole, James and Forsyth, David A},
  booktitle={Proceedings of the IEEE/CVF Conference on Computer Vision and Pattern Recognition},
  pages={4231--4240},
  year={2024}
}

@inproceedings{xing2025luminet,
  title={Luminet: Latent intrinsics meets diffusion models for indoor scene relighting},
  author={Xing, Xiaoyan and Groh, Konrad and Karaoglu, Sezer and Gevers, Theo and Bhattad, Anand},
  booktitle={Proceedings of the Computer Vision and Pattern Recognition Conference},
  pages={442--452},
  year={2025}
}

@article{zhang2024latent,
  title={Latent intrinsics emerge from training to relight},
  author={Zhang, Xiao and Gao, William and Jain, Seemandhar and Maire, Michael and Forsyth, David and Bhattad, Anand},
  journal={Advances in Neural Information Processing Systems},
  volume={37},
  pages={96775--96796},
  year={2024}
}

@inproceedings{taniai2018neural,
  title={Neural inverse rendering for general reflectance photometric stereo},
  author={Taniai, Tatsunori and Maehara, Takanori},
  booktitle={International Conference on Machine Learning},
  pages={4857--4866},
  year={2018},
  organization={PMLR}
}

@inproceedings{zhu2022learning,
  title={Learning-based inverse rendering of complex indoor scenes with differentiable monte carlo raytracing},
  author={Zhu, Jingsen and Luan, Fujun and Huo, Yuchi and Lin, Zihao and Zhong, Zhihua and Xi, Dianbing and Wang, Rui and Bao, Hujun and Zheng, Jiaxiang and Tang, Rui},
  booktitle={SIGGRAPH Asia 2022 Conference Papers},
  pages={1--8},
  year={2022}
}

@inproceedings{zeng2024rgb,
author = {Zeng, Zheng and Deschaintre, Valentin and Georgiev, Iliyan and Hold-Geoffroy, Yannick and Hu, Yiwei and Luan, Fujun and Yan, Ling-Qi and Ha\v{s}an, Milo\v{s}},
title = {RGB$\leftrightarrow$X: Image decomposition and synthesis using material- and lighting-aware diffusion models},
year = {2024},
isbn = {9798400705250},
publisher = {Association for Computing Machinery},
address = {New York, NY, USA},
url = {https://doi.org/10.1145/3641519.3657445},
doi = {10.1145/3641519.3657445},
booktitle = {ACM SIGGRAPH 2024 Conference Papers},
articleno = {75},
numpages = {11},
location = {Denver, CO, USA},
series = {SIGGRAPH '24}
}

@inproceedings{deng2024flashtex,
  title={Flashtex: Fast relightable mesh texturing with lightcontrolnet},
  author={Deng, Kangle and Omernick, Timothy and Weiss, Alexander and Ramanan, Deva and Zhu, Jun-Yan and Zhou, Tinghui and Agrawala, Maneesh},
  booktitle={European conference on computer vision},
  pages={90--107},
  year={2024},
  organization={Springer}
}

@inproceedings{gao2024relightable,
  title={Relightable 3d gaussians: Realistic point cloud relighting with brdf decomposition and ray tracing},
  author={Gao, Jian and Gu, Chun and Lin, Youtian and Li, Zhihao and Zhu, Hao and Cao, Xun and Zhang, Li and Yao, Yao},
  booktitle={European Conference on Computer Vision},
  pages={73--89},
  year={2024},
  organization={Springer}
}

@article{jin2024neural,
  title={Neural gaffer: Relighting any object via diffusion},
  author={Jin, Haian and Li, Yuan and Luan, Fujun and Xiangli, Yuanbo and Bi, Sai and Zhang, Kai and Xu, Zexiang and Sun, Jin and Snavely, Noah},
  journal={Advances in Neural Information Processing Systems},
  volume={37},
  pages={141129--141152},
  year={2024}
}

@inproceedings{poirier2024diffusion,
  title={A Diffusion Approach to Radiance Field Relighting using Multi-Illumination Synthesis},
  author={Poirier-Ginter, Yohan and Gauthier, Alban and Phillip, Julien and Lalonde, J-F and Drettakis, George},
  booktitle={Computer Graphics Forum},
  volume={43},
  number={4},
  pages={e15147},
  year={2024},
  organization={Wiley Online Library}
}

@inproceedings{kim2024switchlight,
  title={Switchlight: Co-design of physics-driven architecture and pre-training framework for human portrait relighting},
  author={Kim, Hoon and Jang, Minje and Yoon, Wonjun and Lee, Jisoo and Na, Donghyun and Woo, Sanghyun},
  booktitle={Proceedings of the IEEE/CVF Conference on Computer Vision and Pattern Recognition},
  pages={25096--25106},
  year={2024}
}

@inproceedings{hou2022face,
  title={Face relighting with geometrically consistent shadows},
  author={Hou, Andrew and Sarkis, Michel and Bi, Ning and Tong, Yiying and Liu, Xiaoming},
  booktitle={Proceedings of the IEEE/CVF conference on computer vision and pattern recognition},
  pages={4217--4226},
  year={2022}
}

@inproceedings{mei2023lightpainter,
  title={Lightpainter: Interactive portrait relighting with freehand scribble},
  author={Mei, Yiqun and Zhang, He and Zhang, Xuaner and Zhang, Jianming and Shu, Zhixin and Wang, Yilin and Wei, Zijun and Yan, Shi and Jung, HyunJoon and Patel, Vishal M},
  booktitle={Proceedings of the IEEE/CVF conference on computer vision and pattern recognition},
  pages={195--205},
  year={2023}
}

@inproceedings{liu2020learning,
  title={Learning to factorize and relight a city},
  author={Liu, Andrew and Ginosar, Shiry and Zhou, Tinghui and Efros, Alexei A and Snavely, Noah},
  booktitle={European Conference on Computer Vision},
  pages={544--561},
  year={2020},
  organization={Springer}
}

@inproceedings{yu2020self,
  title={Self-supervised outdoor scene relighting},
  author={Yu, Ye and Meka, Abhimitra and Elgharib, Mohamed and Seidel, Hans-Peter and Theobalt, Christian and Smith, William AP},
  booktitle={European Conference on Computer Vision},
  pages={84--101},
  year={2020},
  organization={Springer}
}

@inproceedings{kocsis2024lightit,
  title={Lightit: Illumination modeling and control for diffusion models},
  author={Kocsis, Peter and Philip, Julien and Sunkavalli, Kalyan and Nie{\ss}ner, Matthias and Hold-Geoffroy, Yannick},
  booktitle={Proceedings of the IEEE/CVF Conference on Computer Vision and Pattern Recognition},
  pages={9359--9369},
  year={2024}
}

@article{yu15lsun,
    Author = {Yu, Fisher and Zhang, Yinda and Song, Shuran and Seff, Ari and Xiao, Jianxiong},
    Title = {LSUN: Construction of a Large-scale Image Dataset using Deep Learning with Humans in the Loop},
    Journal = {arXiv preprint arXiv:1506.03365},
    Year = {2015}
}

@inproceedings{li2018learning,
  title={Learning intrinsic image decomposition from watching the world},
  author={Li, Zhengqi and Snavely, Noah},
  booktitle={Proceedings of the IEEE conference on computer vision and pattern recognition},
  pages={9039--9048},
  year={2018}
}

@inproceedings{choi2025scribblelight,
  title={Scribblelight: Single image indoor relighting with scribbles},
  author={Choi, Jun Myeong and Wang, Annie and Peers, Pieter and Bhattad, Anand and Sengupta, Roni},
  booktitle={Proceedings of the Computer Vision and Pattern Recognition Conference},
  pages={5720--5731},
  year={2025}
}

@inproceedings{kajiya1986rendering,
  title={The rendering equation},
  author={Kajiya, James T},
  booktitle={Proceedings of the 13th annual conference on Computer graphics and interactive techniques},
  pages={143--150},
  year={1986}
}

@inproceedings{
    zhang2025scaling,
    title={Scaling In-the-Wild Training for Diffusion-based Illumination Harmonization and Editing by Imposing Consistent Light Transport},
    author={Lvmin Zhang and Anyi Rao and Maneesh Agrawala},
    booktitle={The Thirteenth International Conference on Learning Representations},
    year={2025},
    url={https://openreview.net/forum?id=u1cQYxRI1H}
}

@inproceedings{infinigen2024indoors,
    author    = {Raistrick, Alexander and Mei, Lingjie and Kayan, Karhan and Yan, David and Zuo, Yiming and Han, Beining and Wen, Hongyu and Parakh, Meenal and Alexandropoulos, Stamatis and Lipson, Lahav and Ma, Zeyu and Deng, Jia},
    title     = {Infinigen Indoors: Photorealistic Indoor Scenes using Procedural Generation},
    booktitle = {Proceedings of the IEEE/CVF Conference on Computer Vision and Pattern Recognition (CVPR)},
    month     = {June},
    year      = {2024},
    pages     = {21783-21794}
}

\end{document}